\documentclass[lettersize,journal]{IEEEtran}
\usepackage{graphicx}
\usepackage{amsmath}
\usepackage{amssymb}
\usepackage{booktabs}

\usepackage{algorithm}
\usepackage{algorithmic}
\usepackage{multirow}
\usepackage[table,xcdraw]{xcolor}
\usepackage{amssymb}
\usepackage{bbding}
\usepackage{autobreak}
\usepackage{makecell, amsmath}
\usepackage{bm}
\usepackage{arydshln}
\usepackage{subcaption}

\usepackage[pagebackref,breaklinks,colorlinks]{hyperref}
\usepackage[abs]{overpic}
\usepackage[capitalize]{cleveref}

\usepackage{tabulary}

\newcommand{\ours}[1]{{LaVIDE}}
\definecolor{myblue}{RGB}{0,0,255}

\ifCLASSOPTIONcompsoc
  \usepackage[nocompress]{cite}
\else
  \usepackage{cite}
\fi

\ifCLASSINFOpdf
  
\else
 
\fi

\begin{document}
\title{LaVIDE: Language-Prompted Satellite Change Detection via Map-Image Alignment}

\author{Shuguo Jiang,
        Fang Xu,
        Chuandong Liu,
        Hong Tan,
        Shengyang Li,
        Lei Yu,
        Wen Yang,
        Sen Jia,
        and Gui-Song Xia

\IEEEcompsocitemizethanks{
\IEEEcompsocthanksitem Shuguo Jiang and Chuandong Liu are with the School of Computer Science, Wuhan University, Wuhan, 430072, China (email: shuguoj@foxmail.com, chuandong.liu@whu.edu.cn).
\IEEEcompsocthanksitem Fang Xu, Lei Yu, and Gui-Song Xia are with the School of Artificial Intelligence, Wuhan University, Wuhan, 430072, China (email: xufang@whu.edu.cn, ly.wd@whu.edu.cn, guisong.xia@whu.edu.cn).
\IEEEcompsocthanksitem Hong Tan is with the Technology and Engineering Center for Space Utilization and the Key Laboratory of Space Utilization, Chinese Academy of Sciences, Beijing 100094, China (email: tanhong@csu.ac.cn).
\IEEEcompsocthanksitem Shengyang Li is with the Technology and Engineering Center for Space Utilization and the Key Laboratory of Space Utilization, Chinese Academy of Sciences, Beijing 100094, China, and the School of Aeronautics and Astronautics, University of Chinese Academy of Sciences, Beijing 100049, China (email: shyli@csu.ac.cn).
\IEEEcompsocthanksitem Wen Yang is with the School of Electronic Information, Wuhan University, Wuhan, 430072, China (email: yangwen@whu.edu.cn).
\IEEEcompsocthanksitem Sen Jia is with the College of Computer Science and Software Engineering, Shenzhen University, Shenzhen, 518060, China (email: senjia@szu.edu.cn).
}
\thanks{(Corresponding author: Fang Xu, Hong Tan, and Gui-Song Xia.)}
}

\markboth{IEEE Transactions on Pattern Analysis and Machine Intelligence}
{IEEE Transactions on Pattern Analysis and Machine Intelligence}
\maketitle
\begin{abstract}
Remote sensing change detection based on a map reference and an up-to-date image boosts timely observation of the Earth's surface when earlier images are lacking for comparison.
However, the semantic gap between high-level map categories and low-level image details hinders the extraction of homogeneous features for robust temporal association in change detection.
Unlike conventional approaches that either compare pixel-level visual similarity or propagate segmentation errors, \textcolor{black}{we propose a novel framework, \underline{La}nguage-\underline{VI}sion \underline{D}iscriminator
for d\underline{E}tecting changes, LaVIDE}, which bridges the semantic gap between high-level map categories and low-level image details using language as an intermediary.
Specifically, we introduce {\it restricted prompt learning} to generate context-aware textual prompts that align map semantics with image content, and an {\it object-aware embedding enhancement} strategy to integrate object-level attributes (e.g., shape, boundary) into map representations. These components enable robust cross-modal alignment within a unified language-vision feature space. Extensive experiments on four benchmarks, DynamicEarthNet, HRSCD, BANDON, and SECOND, demonstrate that LaVIDE outperforms state-of-the-art methods by significant margins, achieving $18.4\%$ and $5.2\%$ improvements in IoU on multi-class and single-class change detection tasks, respectively. Our framework not only advances the accuracy of map-image change detection but also provides a practical solution for rapid map updating with minimal human intervention, promising broad impacts in urban planning, disaster assessment, and ecological conservation. Code and datasets are available at: \url{https://github.com/ShuGuoJ/LAVIDE.git}.

\end{abstract}


\section{Introduction}\label{sec:introduction}
\IEEEPARstart{D}{etecting} geospatial changes using a single remote sensing image and an outdated map~\cite{mapformer2023}, i.e., map–image change detection, enables timely perception of temporal dynamics on the Earth's surface. It connects dynamic observations with static geospatial knowledge, \textcolor{black}{thereby enabling direct support for geospatial database maintenance and} facilitating urban monitoring, map updates, among others~\cite{dualtasklearning2022}.

Most existing studies~\cite{changegan2023,change22024} detect surface changes by comparing bi-temporal remote sensing images using low-level visual cues such as texture and color. 
These methods rely on the availability of historical imagery and are generally incapable of handling change detection with map–image pairs.
Unlike homogeneous image pairs, the data in map–image pairs are heterogeneous: \textit{vector-based maps contain objects annotated with explicit geometric and semantic attributes, while raster images offer abundant visual details.}
Thus, existing change detection methods~\cite{star2021,saan2024}, designed for raster-encoded images, are ill-equipped to interpret or exploit the high-level semantics embedded in maps.
Furthermore, the high-level categorical information encoded in maps contrasts
with the low-level visual information in images.  Bridging this
information gap poses a significant
challenge in 
accurately detecting changes in map-image pairs.

\begin{figure}[t]
    \centering
    \includegraphics[width=0.98\columnwidth]{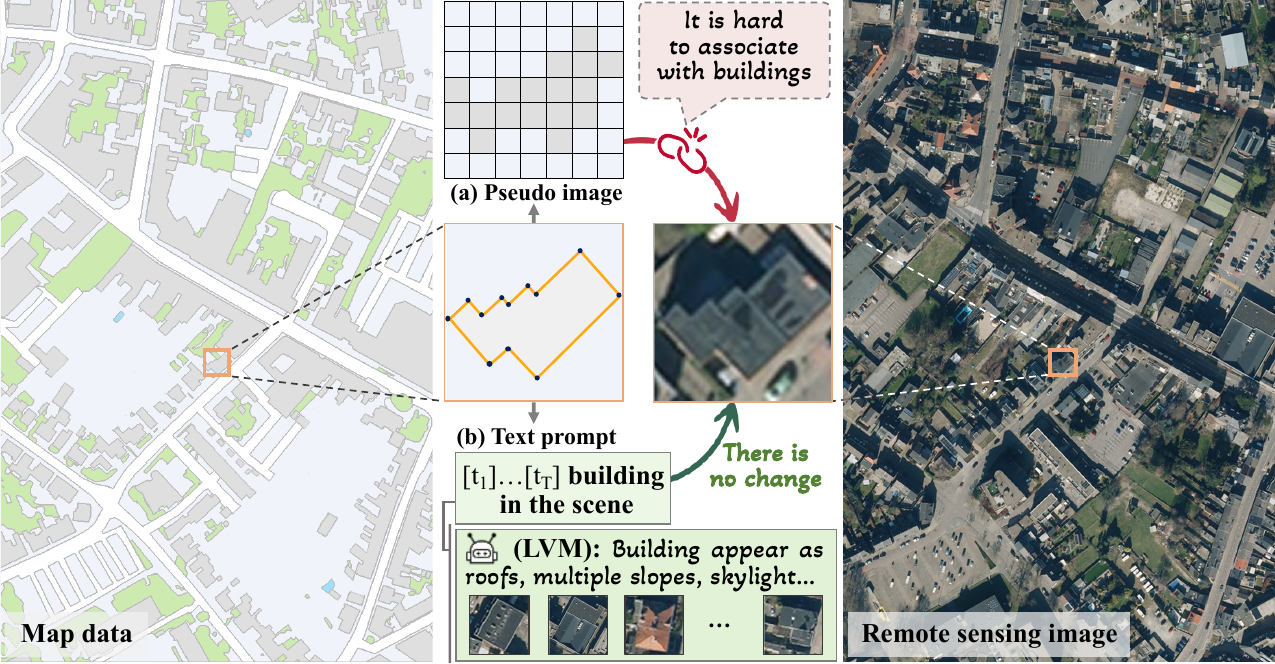}
    \caption{\textcolor{black}{Motivation illustration. Unlike existing works~\cite{mapformer2023, objecttransformer2023, samcd2024}, which convert maps into pseudo-images and rely on visual comparison to detect changes (a), \ours{} leverages language as an intermediary to bridge the semantic gap between high-level map categories and low-level image details (b).}}
    \label{fig:language_verus_color}
    \vspace{-6mm}
\end{figure}

To date, few explorations have been dedicated to change detection via the comparison of maps and remote sensing images. The existing works~\cite{mapformer2023, objecttransformer2023, samcd2024} convert maps into pseudo-images by assigning unique colors to different semantic categories, \textcolor{black}{as shown in~\cref{fig:language_verus_color}(a)}, and then rely on visual comparison to identify change regions.  
However, this process degrades the high-level semantic priors inherently embedded in maps into purely low-level visual representations, which lack sufficient discriminative power for meaningful semantic comparison.
Moreover, aligning color-based pseudo-visual features with the complex textures in remote sensing images remains highly difficult, hindering reliable cross-modal semantic association.
In this paper, rather than making maps ``look like images”, we concentrate on explicitly modeling their high-level semantics for more reliable and interpretable alignment with remote sensing imagery.

Language encodes high-level semantics grounded in extensive world knowledge. Unlike purely visual cues, language-based representations inherently capture semantic priors, such as \textit{how buildings typically appear in remote sensing imagery}, thereby linking symbolic knowledge with visual perception, \textcolor{black}{as shown in~\cref{fig:language_verus_color}(b)}.
So,
we propose to prompt ground objects in maps using language and leverage the semantic association between language and vision to align maps with images, thus tackling the challenges of map–image change detection.
We present a novel framework, \textit{\textbf{La}nguage–\textbf{VI}sion \textbf{D}iscriminator for d\textbf{E}tecting changes using an outdated map and a single image}, termed LaVIDE, \textcolor{black}{which discriminates whether regions in maps have changed or remained unchanged by encoding} map semantics into natural language and \textcolor{black}{aligning} them with satellite imagery within a unified language–vision feature space.
Maps are transformed into textual prompts describing the semantic categories and attributes of ground objects, which are then processed by a language–vision model (LVM), e.g., CLIP~\cite{clip2021}, to fully exploit map semantics. 
By aligning these with the visual patterns in the image,  \ours{} enables semantically grounded change detection beyond low-level visual comparison.

Noticing that remote sensing images are inherently scene-centric, where the appearance of ground objects varies significantly with geographic context and land-use patterns, we propose a \textit{restricted prompt learning} approach that automatically generates context-aware textual prompts to more effectively align symbolic map semantics with the complex visual content of remote sensing imagery.
We further introduce an \textit{object-aware embedding enhancement} strategy that integrates object-level attributes into the map embeddings, enabling more precise comparison with image details.
On the image side, \ours{} distills knowledge from the LVM into a task-specific hierarchical vision encoder, enriching multi-scale image representations while preserving language–vision associations for robust change detection.

To sum up, the contributions of this work are three-fold:
\begin{itemize}
\item We propose a novel map-image change detection network, \ours{}, which introduces language to bridge the high-level semantic categories from maps and the low-level visual details from images,  thus boosting change detection performance.
 
\item We propose a restricted prompt learning approach that efficiently generates prompts consistent with image content for map representations, and refines them with object attributes at feature level, thereby enhancing semantic alignment with image details.

\item Extensive experiments demonstrate that the proposed \ours{} significantly improves performance in land use change detection (i.e., multi-class) and building change detection (i.e., single-class), achieving IoU improvements of 18.4\% on DynamicEarthNet~\cite{dynamicearthnet2022}, 5.2\% on HRSCD~\cite{hrscd2019}, 3.6\% on BANDON~\cite{bandon2023}, and 1.6\% on SECOND~\cite{second2022}.
\end{itemize}

\vspace{-2mm}
\section{Related Work}
\label{sec: related work}
\subsection{Bi-Temporal Change Detection}
Bi-temporal change detection\textcolor{black}{~\cite{cdmamba2025,change3d2025}} refers to the process of identifying changes between two satellite images captured at different times over the same geographical area.
Early research~\cite{hsicd2014} in this field relies heavily on human expertise for hand-crafted feature engineering to identify pixel differences. 
They are highly susceptible to variations in illumination and seasonal conditions, often misinterpreting differences caused by external factors as actual changes.
In recent years, learning-based methods\textcolor{black}{~\cite{clafa2023,bit2021}} have shown great progress in detecting changes owing to their powerful feature extraction capabilities.
They typically assume that images captured at different times follow the same distribution, thereby using siamese network\textcolor{black}{~\cite{isnet2022}},~\cite{saan2024}, which shares weights between two branches but with identical architecture, to process bi-temporal images. 
However, the heterogeneous nature of maps and images results in non-identical data distributions, which undermines the effectiveness of bi-temporal change detection methods in extracting homogeneous features for reliable map-image change detection.
\vspace{-3mm}
\subsection{Cross-Modal Change Detection} 
Cross-modal change detection reveals surface changes using images from different times and modalities, relying on the homogenization of cross-modal information to enable semantic comparison.
In the field of cross-modal change detection, comparing maps and images plays a crucial role in enabling timely change detection.
To homogenize maps and images, 
some works~\cite{buildingsm2018,osmsm2017} transform images into map-like data through semantic recognition techniques, e.g., semantic segmentation, allowing for direct comparison with maps.
The effectiveness of change detection in these methods is constrained by the accuracy of the transformation process.
To address these issues, some works~\cite{mapformer2023,objecttransformer2023} transform maps into image-like data by using color or one-hot encoding to represent ground objects, facilitating change detection through cross-modal methods.
By jointly training the entire change detection process under supervised learning, these methods better integrate feature extraction and comparison, achieving more accurate results.
However, cross-modal change detection methods based on visual comparison are often affected by the intrinsic visual discrepancies between maps and images, hindering progress in the field.

\begin{figure*}[!t]
    \centering
    \includegraphics[width=0.85\textwidth]{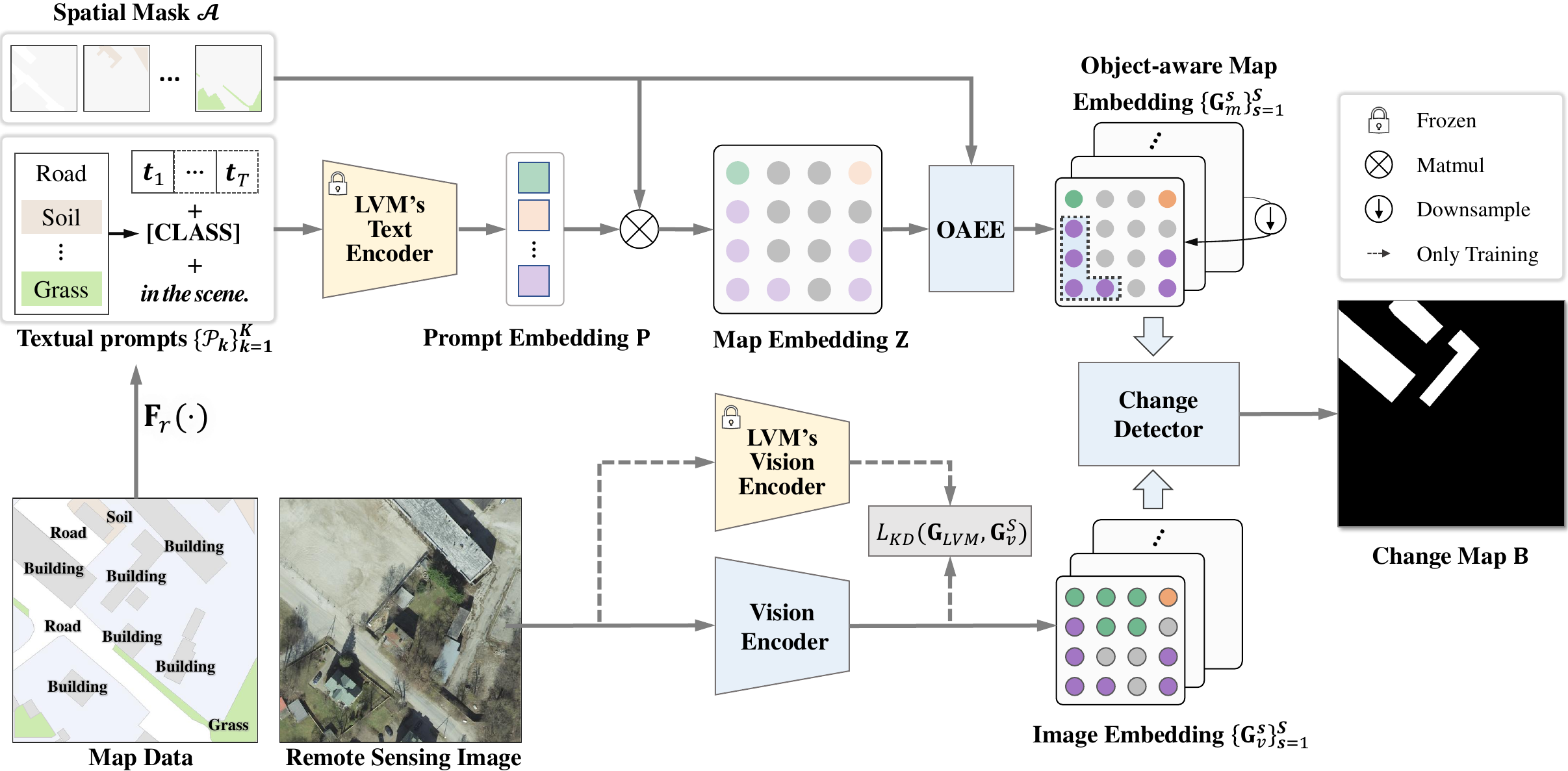}
    \caption{\textcolor{black}{The pipeline of our proposed \ours{}, which leverages language to bridge the information gap between maps and images. OAEE denotes the Object-Aware Embedding Enhancement module.}}
    \label{fig:architecture}
    \vspace{-6.5mm}
\end{figure*}
\vspace{-2mm}
\section{Methodology}
\label{sec:Method}
\subsection{Problem Statement}
Given a satellite image $\mathbf{I}\in \mathbb{R}^{H\times W\times 3}$ and an outdated map $\mathcal{M}$, the objective of change detection is to identify regions occurring semantic replacement between them.
Here, $H$ and $W$ denote the image height and width, respectively.
The identified regions are denoted by a binary change map $\mathbf{B} \in \{0,1\}^{H \times W}$, where 0 indicates no change and 1 indicates change.
This task is generally approached through a homogenization process that transforms heterogeneous inputs into a consistent representation, followed by a change detection step that compares the homogenized data to identify changes.
The change detection process is defined as follows:
\begin{equation}
    \mathbf{B} = \text{H}(\text{F}^{\mathcal{M}}(\gamma (\mathcal{M})), \text{F}^{\mathbf{I}}(\mathbf{I})),
\label{eq: change detection}
\end{equation}
where $\gamma(\cdot)$ denotes a conversion operator that transforms the map $\mathcal{M}$ into a rasterized representation suitable for subsequent processing, $\text{F}^{{\mathcal{M}/\mathbf{I}}}(\cdot)$ are modality-specific backbones that encode the map and image into a unified latent space, and $\text{H}(\cdot)$ is a change detection function that compares the aligned representations to produce the final change map $\mathbf{B}$.

\noindent{\textit{Category discrimination}}:  
The basic strategy involves inferring a semantic map from the satellite image to align with the map, enabling a direct comparison.
Specifically, the map $\mathcal{M}$ is first converted by the operator $\gamma(\cdot)$ into a semantic raster $\gamma(\mathcal{M}) \in \{1, \dots, K\}^{H \times W}$, where each pixel is assigned a categorical label representing a ground object class, and $K$ denotes the number of semantic categories. 
The map encoder $\text{F}^{\mathcal{M}}(\cdot)$ is implemented as an identity function, i.e., $\text{F}^{\mathcal{M}}(\gamma(\mathcal{M})) = \gamma(\mathcal{M})$, while the image encoder $\text{F}^{\mathbf{I}}(\cdot)$ is realized by a semantic segmentation network $\text{Seg}(\cdot)$ that predicts a dense category label map from the satellite image: $\text{F}^{\mathbf{I}}(\mathbf{I}) = \text{Seg}(\mathbf{I}) \in \{1, \dots, K\}^{H \times W}$.
The comparison module $\text{H}(\cdot)$ then performs label-wise matching to identify changes. 
The effectiveness of this approach is critically dependent on the accuracy of the semantic map generated by $\text{Seg}(\cdot)$. Segmentation errors, including misclassifications and boundary inaccuracies, can substantially compromise the accuracy of the change detection process.

\noindent{\textit{Vision discrimination} }:
By converting the map into a visually comparable format, cross-modal change detection techniques can be employed to assess differences directly.
Specifically, the converter $\gamma(\cdot)$ typically transforms the map $\mathcal{M}$ into a pseudo-RGB image $\gamma(\mathcal{M}) \in \mathbb{R}^{H \times W \times 3}$ by assigning unique colors to semantic categories. The pseudo-image and the satellite image $\mathbf{I}$ are then independently encoded using vision-specific backbones $\text{F}^{\mathcal{M}}(\cdot)$ and $\text{F}^{\mathbf{I}}(\cdot)$ (e.g., CNNs or vision transformers), yielding feature maps in a shared visual space. 
The comparison function $\text{H}(\cdot)$, typically implemented as a lightweight decoder, is used to identify changes by comparing these features.
However, using customized colors to indicate categories fails to reflect the realistic characteristics of ground objects, thereby weakening the semantic expressiveness of the map. Moreover, it introduces a domain gap that increases the difficulty of semantic discrimination.

Thus, the main obstacles to boosting change detection performance are two-fold: 
\begin{itemize}
    \item The map encoding should preserve high-level semantic information of ground objects to facilitate effective feature extraction.
    \item The model should establish meaningful associations between map categories and image details to minimize the semantic gap between maps and satellite imagery.
\end{itemize}

\noindent{\textbf{\textit{Language-vision discrimination}}}:
Thus, the task of map-image change detection is to develop a map converter that effectively retains high-level categorical information, facilitating semantic feature extraction.
Meanwhile, it necessitates the design of a model that associates map categories with image details, mitigating the cross-modal discrepancy.
To this end, we define the converter $\gamma(\cdot)$ to generate a language-format map representation $\mathcal{T} = (\mathcal{C}, \bm{\mathcal{A}})$.
The $\mathcal{C}$ denotes a text set that carries the semantic information of ground objects (e.g., $\mathcal{C}_k=\text{``building''}$), and $\bm{\mathcal{A}}\in\mathbb{R}^{K\times H\times W}$ is a spatial mask matrix, where $\bm{\mathcal{A}}_{k}\in \{0,1\}^{H\times W}$ indicates the spatial distribution of $\mathcal{C}_{k}$.
Since linguistic symbols inherently encapsulate the semantic characteristics of ground objects, the resulting map encoding can effectively convey high-level information. 
Moreover, the intrinsic connection between language and visual perception enables semantic alignment between the map and image.
Specifically, the map encoder $\text{F}^{\mathcal{M}}(\cdot)$ is implemented as a text encoder that processes the linguistic representations of the map, while the image encoder $\text{F}^{\mathbf{I}}(\cdot)$ is implemented as a vision encoder that extracts visual features from satellite imagery.
Therefore, the change detection operator $\text{H}(\cdot)$, which operates on linguistic features from the map and visual features from the image, is capable of bridging the semantic gap.
\vspace{-4mm}
\subsection{Overall}
The overall framework of the proposed \ours{} algorithm is illustrated in ~\cref{fig:architecture}. It aligns
maps and images within the feature space of a language-vision model, enabling the association of high-level semantic categories from maps with low-level visual details from images. 
\ours{} comprises two branches, the map branch and the image branch, responsible for extracting multi-scale map embeddings \textcolor{black}{$\{\mathbf{G}_{m}^{s}\}_{s=1}^{S}$} and image embeddings \textcolor{black}{$\{\mathbf{G}_{v}^{s}\}_{s=1}^{S}$}, respectively, \textcolor{black}{where $S$ denotes the number of scales, set to 4 in this work.}

Concretely, to enrich ground objects' semantic information, the map branch obtains text prompt \textcolor{black}{$\mathcal{P}_k$} for each category \textcolor{black}{$\mathcal{C}_k$} through a restricted object prompt function \textcolor{black}{$\text{F}_{r}(\cdot)$}, i.e., $\mathcal{P}_{k}=\text{F}_{r}(\mathcal{C}_{k})$.
The prompt embedding $\mathbf{P}$ is obtained by feeding prompts \textcolor{black}{$\{\mathcal{P}_{k}\}_{k=1}^{K}$ into the text encoder of the language–vision model, where $K$ denotes the number of categories.} 
\textcolor{black}{$\mathbf{P}$ is multiplied with the spatial mask $\bm{\mathcal{A}}$ to generate the map embedding $\mathbf{Z}$} in the unified language–vision feature space.
To further align map semantics with image content, the map branch incorporates object-level features (e.g., shape and boundary) \textcolor{black}{into the map embedding $\mathbf{Z}$ through an Object-Aware Embedding Enhancement (OAEE) module, producing the object-aware map embedding $\{\mathbf{G}_{m}^{s}\}_{s=1}^{S}$}.
The image branch includes a hierarchical vision encoder $\text{F}_{v}(\mathcal{\cdot})$ for multi-scale feature extraction, obtaining the image embeddings \textcolor{black}{$\{\mathbf{G}_{v}^{s}\}_{s=1}^{S}$}.
To ensure semantic alignment between image and map embeddings, we distill knowledge of the language-vision model into the vision encoder. 
The embeddings \textcolor{black}{$\{\mathbf{G}_{m}^{s}\}_{s=1}^{S}$} and \textcolor{black}{$\{\mathbf{G}_{v}^{s}\}_{s=1}^{S}$} are then fed into a change detector to identify differences.
\vspace{-4mm}
\subsection{Map-image Feature Extraction}
\subsubsection{Map Branch}
We use the text encoder of the language-vision foundation model to extract high-level semantic features of maps from prompts constructed with category-specific text.
As the text encoder is trained jointly with the visual encoder via image-text alignment objectives, the extracted map features are semantically relevant to image content.
Given the sensitivity of the text encoder to prompt content and cumbersome manual prompt design, we propose a restricted prompt learning approach.
It integrates the characteristics of task gradients and remote sensing images to jointly optimize prompts tailored for encoding maps, enabling the resulting map embeddings to semantically align with image embeddings.

\textcolor{black}{\textit{Restricted Prompt Learning:}} Considering that task gradients are effective in guiding the design of task-oriented prompts, we introduce $T$ learnable tokens as contextual prefixes to each category text. 
The learnable tokens receive task information to generate appropriate semantic context for ground objects, thereby promoting better feature alignment.
Furthermore, since remote sensing images are scene-centric, unlike natural images, which typically focus on a single object, we explicitly embed prompts within the contextual phrase ``in the scene''.
The contextual phrase serves as a constraint on prompt learning and better captures the spatial semantics of remote sensing imagery.
As a result, the final prompt $\mathcal{P}_{k}$ used for map encoding takes the form of ``[learnable tokens]+[$C_{k}$]+[in the scene]'', enabling the representation of map categories to be more consistent with remote sensing imagery by leveraging task-driven gradients and image-specific characteristics.

\textit{Object-Aware Embedding Enhancement:} In addition to categorical attributes, map data includes specific object information, e.g., shape and boundary features. To further strengthen map information with respect to object attributes, \textcolor{black}{the map branch uses an object-aware embedding enhancement module to incorporate object-specific attribute features into map embedding. 
Concretely, for each category of ground objects $\mathcal{C}_{k}$, we extract its spatial mask from the map, denoted as $\bm{\mathcal{A}}_{k}$. 
An object encoder $\text{F}_{obj}(\cdot)$ is then used to extract object-specific attribute features $\mathbf{O}$ from the spatial masks $\bm{\mathcal{A}}=\{\bm{\mathcal{A}}_{k}\}_{k=1}^{K}$, i.e., $\mathbf{O}=\text{F}_{obj}(\bm{\mathcal{A}})$.
The extracted feature is concatenated with the map embedding $\mathbf{Z}$ and processed by a convolutional network with three convolutional modules, whose output is added to $\mathbf{Z}$ to obtain the object-aware map embedding $\mathbf{G}_{m}^{S}$. To align the spatial scales of the map and image embeddings, we generate multi-scale object-aware map embeddings $\{\mathbf{G}_{m}^{s}\}_{s=1}^{S}$ from $\mathbf{G}_{m}^{S}$ via a simple interpolation operation. More details on OAEE are provided in Sec. S1.A of supplementary material.}

\subsubsection{Image Branch}
An intuitive way to obtain image embeddings that are semantically aligned with text is to use the vision encoder of the language-vision model. However, the language-vision models' encoder, e.g., CLIP, is typically built on a flat architecture, lacking the ability to extract multi-scale features, which is essential for detecting changes at various scales. 
Multi-scale changes are inherent to remote-sensing change detection because ground objects, whether belonging to the same or to different categories, often vary greatly in scale.

To this end, we adopt a hierarchical architecture to extract multi-scale image embeddings \textcolor{black}{$\{\mathbf{G}_{v}^{s}\}_{s=1}^{S}$}. Specifically, we employ the feature backbone of SegFormer \cite{segformer2021} as the vision encoder, generating features at four different scales.
To establish semantic consistency between visual and map embeddings, we adopt a feature distillation strategy that aligns the feature space of the hierarchical vision encoder with that of the vision encoder of the language-vision model. 
Owing to architectural heterogeneity between the two encoders, a feature mismatch arises during the distillation process.

To alleviate this, we supervise only the final-layer image embedding $\mathbf{G}_{v}^{S}$ from the hierarchical vision encoder using the output feature $\mathbf{G}_{LVM}$ from the vision encoder of the language-vision model, thereby simplifying feature alignment between the two structurally distinct encoders.
Moreover, considering the potential conflicts between general language-vision knowledge and task-specific change detection semantics, we adopt a correlation loss rather than a consistency loss as the distillation objective $L_{KD}$. 
\textcolor{black}{Unlike the consistency loss (e.g., MSE~\cite{distilling2023}), which enforces element-wise numerical matching, the correlation loss aligns the relationships between feature representations, enabling the hierarchical vision encoder to absorb semantic knowledge from the language–vision model while preserving discriminative representations for change detection.} The correlation loss is calculated as:
\begin{equation}
    \textcolor{black}{L_{\text{KD}}= \frac{1}{H_{S}W_{S}}\sum _{i=1}^{H_{S}}\sum _{j=1}^{W_{S}}\big(1-\cos{((\mathbf{G}_{v}^{S})_{ij}}, (\mathbf{G}_{\text{LVM}})_{ij})\big),}
\end{equation}
in which $\cos(\cdot)$ denotes the cosine similarity function, \textcolor{black}{and $H_{S}$ and $W_{S}$ denote the height and width of the $S$-th scale embedding, respectively.}
\vspace{-2mm}
\subsection{Map-image Change \textcolor{black}{Detector}}
The detectability of changes is heavily influenced by inter-category similarity. 
Changes between highly similar classes, such as vegetation and agricultural fields, are more subtle than those between dissimilar classes, exemplified by vegetation and buildings.
To effectively capture such subtle changes, discriminative features should comprehensively consider diverse semantic differences from multiple perspectives, enabling robust change detection.
Inspired by the way humans compare objects across multiple semantic dimensions, such as texture and color, and adaptively focus on salient differences to determine equivalence, we implement the change detector with a MoE discriminative module.

The MoE discriminative module employs $N$ experts \textcolor{black}{$\{\text{E}_{n}(\cdot)\}_{n=1}^{N}$} to measure object differences from diverse semantic perspectives, each of which is achieved by a multilayer perceptron (MLP). To adaptively focus on significant differences, the MoE discriminative module comprises a change-specific route function $\text{F}_{route}(\cdot)$, which is implemented by a depthwise separable convolutional module, to adaptively assign weights to each perspective.
Specifically, the semantic difference for the $n$-th perspective is quantified as \textcolor{black}{$\mathbf{D}_{n}^{s} = \text{E}_{n}(\mathbf{G}_{m}^{s} \parallel \mathbf{G}_{v}^{s})$}.
The corresponding weight, which reflects the importance of the $n$-th semantic perspective, is calculated as \textcolor{black}{$\mathbf{W}_{n}^{s} = \text{F}_{route}(\mathbf{G}_{m}^{s} \parallel \mathbf{G}_{v}^{s})$}. 
The final discriminative feature for the $s$-th scale is then calculated by weighting \textcolor{black}{$\mathbf{D}_{n}^{s}$} with \textcolor{black}{$\mathbf{W}_{n}^{s}$}, i.e., \textcolor{black}{$\mathbf{D}^{s} = \sum_{n=1}^{N} \mathbf{W}_{n}^{s}\mathbf{D}_{n}^{s}$}. 
These scale-specific discriminative features \textcolor{black}{$\{\mathbf{D}^{s}\}_{s=1}^{S}$} are linearly fused via an MLP into a unified multi-scale feature $\mathbf{D}$ , which is subsequently fed into a binary classifier to generate the final change detection prediction $\mathbf{B}$.
\vspace{-2mm}
\subsection{Loss Function}
We adopt a composite loss to supervise the model training. The primary objective is a cross-entropy loss for binary change detection, defined as:
\begin{equation}
    L_{CD} = \text{CrossEntropy}(\mathbf{B}, \mathbf{GT}),
\end{equation}
where $\mathbf{GT}$ denotes the ground-truth binary change map.
To further encourage the model to learn discriminative features, a contrastive loss $L_{CL}$ is used to push unchanged features in map-image pairs together while pulling changed features apart. 
First, the multi-scale image embeddings \textcolor{black}{$\{\mathbf{G}_{v}^{s}\}_{s=1}^{S}$} are linearly fused via an MLP to obtain a compact feature representation $\mathbf{G}_{CL}$, which is then contrastive with the $S$-th scale object-aware map embedding $\mathbf{G}_{m}^{S}$ as follows:
\begin{equation}
l_{ij}=
\begin{cases}
-\text{cos}\big((\mathbf{G}_{CL})_{ij}, (\mathbf{G}_{m}^{S})_{ij}\big) &\mathbf{GT}_{ij}=0\\
max\big(\text{cos}((\mathbf{G}_{CL})_{ij}, (\mathbf{G}_{m}^{S})_{ij}), 0) & \mathbf{GT}_{ij}=1
\end{cases},
\end{equation}
where $max(\cdot)$ refers to the maximum function.
Thus, the contrastive loss is calculated as:
\begin{equation}
    \textcolor{black}{L_{CL} = \frac{1}{H_{S}W_{S}}\sum _{i=1}^{H_{S}} \sum _{j=1}^{W_{S}} l_{ij}.}
\end{equation}
The overall loss function is as follows: 
\begin{equation}
    L = L_{CD} + \lambda _{1} L_{KD} + \lambda _{2} L_{CL},
\label{eq:loss}
\end{equation}
where $\lambda _{1}$ and $\lambda _{2}$ denote the balancing parameters.

\section{Experiments}
\label{sec:Experiments}

\begin{table*}[!t]
    \setlength{\tabcolsep}{4pt}
    \centering
    \caption{Quantitative comparisons of proposed \ours{} to state-of-the-art methods, with the best-performing results indicated in bold and the next-best results underlined. The results of UPerNet, SETR\_PUP, SegFormer, and MapFormer on DynamicEarthNet and HRSCD stem from \cite{mapformer2023}, while the results of the other methods are reproduced in this study. All results on the SECOND dataset are obtained using models trained on BANDON.
    }
    \label{tab:experiments on dynamicearthnet and hrscd}
    \begin{tabular}{lc>{\centering\arraybackslash} p{0.9cm}>{\centering\arraybackslash} p{0.9cm}>{\centering\arraybackslash} p{0.9cm}>{\centering\arraybackslash} p{0.9cm}>{\centering\arraybackslash} p{0.9cm}>{\centering\arraybackslash} p{0.9cm}>{\centering\arraybackslash} p{0.9cm}>{\centering\arraybackslash} p{0.9cm}>{\centering\arraybackslash} p{1.2cm}>{\centering\arraybackslash} p{1.2cm}}
        \toprule
        \makecell[c]{\multirow{2}{*}{Methods}} & \multirow{2}{*}{Type} & \multicolumn{2}{c}{DynamicEarthNet} & \multicolumn{2}{c}{HRSCD} & \multicolumn{2}{c}{BANDON} & \multicolumn{2}{c}{SECOND} & \multirow{2}{*}{\textcolor{black}{Params.(M)}} & \multirow{2}{*}{\textcolor{black}{FLOPs(G)}}\\ \cmidrule(lr){3-4}\cmidrule(lr){5-6}\cmidrule(lr){7-8}\cmidrule(lr){9-10}
         & & F1(\%) & IoU(\%) & F1(\%) & IoU(\%) & F1(\%) & IoU(\%) & F1(\%) & IoU(\%)\\ \midrule
        UPerNet \cite{upernet2018}& Category & 21.8  & 12.2 & 5.6  & 2.8 & 22.7  & 12.8 & 48.1 & 31.7 & \textcolor{black}{59.8} & \textcolor{black}{442.9} \\
        SETR\_PUP \cite{setrpup2021}& Category & 20.8  & 11.6 & 5.4  & 2.8 & 24.8 & 14.2 & 51.2 & 34.4 & \textcolor{black}{307.8} & \textcolor{black}{115.7} \\
        SSG2 \cite{ssg22024}& Category & 17.2  & 9.4 & 3.7  & 1.8 & 23.7 & 13.4 & 47.9 & 31.5 & \textcolor{black}{72.4} & \textcolor{black}{569.5}\\
        SegFormer \cite{segformer2021}& Category & 21.2  & 11.9 & 5.6  & 2.8 & 25.1 & 14.4 & 51.2 & 34.4  & \textcolor{black}{27.5} & \textcolor{black}{37.7} \\  \midrule
        SNUNet \cite{snunet2021}& Vision & 4.4  & 2.3  & 49.8  & 33.2 & 42.0 & 26.6 & 39.6 & 24.7 & \textcolor{black}{3.3} & \textcolor{black}{110.6} \\
        CGNet \cite{cgnet2023}& Vision & 16.0 & 8.7  & \underline{62.9} & \underline{45.9}  & 71.3 & 55.4  & 65.0 & 48.1 & \textcolor{black}{48.4} & \textcolor{black}{660.5} \\
        ChangeFormer \cite{changeformer2022}& Vision & 31.2  & 18.5 & 62.1  & 45.0 & 69.8 & 53.6 & 65.1 & 48.3 & \textcolor{black}{41.1} & \textcolor{black}{1590.0} \\
        FHD \cite{fhd2022}& Vision & 32.8  & 19.6 & 56.4  & 39.3 & 69.0 & 52.7 & 64.2 & 47.3 & \textcolor{black}{29.2} & \textcolor{black}{118.1} \\
        ChangerEx \cite{changer2023}& Vision & 15.3  & 8.3 & 20.6  & 11.5 & 26.3 & 15.2 & 32.1 & 19.1 & \textcolor{black}{26.6} & \textcolor{black}{75.4} \\ 
        \textcolor{black}{ChangeMamba~\cite{changemamba2024}} & \textcolor{black}{Vision} & \textcolor{black}{32.5} & \textcolor{black}{19.4} & \textcolor{black}{53.4} & \textcolor{black}{36.5} & \textcolor{black}{55.5} & \textcolor{black}{38.4} & \textcolor{black}{50.4} & \textcolor{black}{33.7} & \textcolor{black}{91.6} & \textcolor{black}{209.5}\\
        \textcolor{black}{CDMamba~\cite{cdmamba2025}} & \textcolor{black}{Vision} & \textcolor{black}{30.0} & \textcolor{black}{17.7} & \textcolor{black}{53.0} & \textcolor{black}{36.0} & \textcolor{black}{48.7} & \textcolor{black}{32.2} & \textcolor{black}{42.0} & \textcolor{black}{26.6} & \textcolor{black}{19.2} & \textcolor{black}{243.0} \\
        MapFormer \cite{mapformer2023}& Vision & 32.0  & 19.0 & 62.1  & 45.0 & 71.4  & 55.5 & \underline{67.3} & 50.7 & \textcolor{black}{34.5} & \textcolor{black}{90.1} \\ \midrule
        \ours{} (CLIP) & Language-Vision & \underline{37.6}  & \underline{23.2} & \textbf{65.1}  & \textbf{48.3} & \textbf{73.0} & \textbf{57.5} & \textbf{68.1} & \underline{51.5} & \textcolor{black}{59.5} & \textcolor{black}{389.9} \\ 
        \textcolor{black}{\ours{} (GeoRSCLIP)} & \textcolor{black}{Language-Vision} & \textcolor{black}{\textbf{37.7}} & \textcolor{black}{\textbf{23.3}} & \textcolor{black}{\textbf{65.1}}  & \textcolor{black}{\textbf{48.3}} & \textcolor{black}{\underline{72.7}} & \textcolor{black}{\underline{57.1}} & \textcolor{black}{\textbf{68.1}} & \textcolor{black}{\textbf{51.7}} & \textcolor{black}{59.5} & \textcolor{black}{389.9} \\ \bottomrule
        \end{tabular}
        \vspace{-0.4cm}
\end{table*}

\begin{figure*}[!t]
    \renewcommand{\tabcolsep}{0.4mm}
    \renewcommand{\arraystretch}{0.50}
    \newcommand{\firstimage}[1]{dynamic_scene_2_#1}
    \newcommand{\secondimage}[1]{bandon_scene_2_#1}
    \centering
    \begin{tabular}{ccccccc}
	
        \includegraphics[width=0.135\linewidth]{image/exp/\firstimage{map}} &
        \includegraphics[width=0.135\linewidth]{image/exp/\firstimage{image}} &
        \includegraphics[width=0.135\linewidth]{image/exp/\firstimage{gt}} &
        \includegraphics[width=0.135\linewidth]{image/exp/\firstimage{segformer}} &
        \includegraphics[width=0.135\linewidth]{image/exp/\firstimage{changeformer}} &
        \includegraphics[width=0.135\linewidth]{image/exp/\firstimage{mapformer}}&
        \includegraphics[width=0.135\linewidth]{image/exp/\firstimage{ours}}
	\\
        \includegraphics[width=0.135\linewidth]{image/exp/\secondimage{map}} &
        \includegraphics[width=0.135\linewidth]{image/exp/\secondimage{image}} &
        \includegraphics[width=0.135\linewidth]{image/exp/\secondimage{gt}} &
        \includegraphics[width=0.135\linewidth]{image/exp/\secondimage{segformer}} &
        \includegraphics[width=0.135\linewidth]{image/exp/\secondimage{changeformer}} &
        \includegraphics[width=0.135\linewidth]{image/exp/\secondimage{mapformer}}&
        \includegraphics[width=0.135\linewidth]{image/exp/\secondimage{ours}}
	\\
	\small{Map} &
        \small{Image} &     
        \small{GT} & 
        \small{\textcolor{black}{SegFormer}} & \small{\textcolor{black}{ChangeFormer}}& \small{MapFormer}&
        \small{LaVIDE}
    \end{tabular}
	
	\caption{\textcolor{black}{Visualization of change detection results from the DynamicEarthNet and BANDON datasets. White corresponds to true positives, black to true negatives, red to false positives, and blue to false negatives.}}
	\vspace{-5mm}
	\label{fig:dynamic earth net comparison}
\end{figure*}

\subsection{Experimental Setup}

\textbf{Datasets and metrics}. Our experiments are conducted on two land use change detection datasets, DynamicEarthNet~\cite{dynamicearthnet2022} and HRSCD~\cite{hrscd2019}, as well as two building change detection datasets, BANDON~\cite{bandon2023} and SECOND~\cite{second2022},
where the semantic labels of pre-change images serve as outdated maps.
Change performance is evaluated using F1-score (F1) and Intersection over Union (IoU).

\textbf{Implementation details}. 
\textcolor{black}{Following the pipeline in Fig.~\ref{fig:architecture}, we implement \ours{} with two language–vision models, i.e., CLIP~\cite{clip2021} and GeoRSCLIP~\cite{rs5m2024}, where CLIP is used by default unless otherwise specified. The models are trained} using PyTorch on two NVIDIA Tesla V100 GPUs.
The AdamW optimizer is used, with a learning rate of $6\times 10^{-5}$,  adjusted by a polynomial decay scheduler with a linear warmup phase.
The batch size is set to $12$ and the maximum number of training iterations is set to $32k$. 
The number of experts $N$ in the MoE discriminative module is set to $10$, and the weighting factors $\lambda _{1}$ and $\lambda _{2}$ are both set to $1$. 
In addition, the number of learnable tokens $T$ for restricted prompt learning is set as $4$. 

\vspace{-2mm}
\subsection{Comparison with State-of-the-art Methods}
We compare the proposed \ours{} with state-of-the-art change detection methods, including category discrimination and vision discrimination approaches. For category discrimination approaches, we follow the setup described in~\cite{mapformer2023}, training semantic segmentation backbones using both pre- and post-change semantic labels. 
The quantitative results are presented in~\cref{tab:experiments on dynamicearthnet and hrscd}. We can find that our proposed \ours{} achieves remarkable improvements over state-of-the-art methods, with IoU gains of 18.4\% on DynamicEarthNet, 5.2\% on HRSCD, 3.6\% on BANDON, and 1.6\% on the out-of-domain dataset SECOND.
We choose two representative scenes, one from the land-use change detection dataset DynamicEarthNet and one from the building-change detection dataset BANDON, to qualitatively evaluate the detection results, as illustrated in~\cref{fig:dynamic earth net comparison}. 
The results obtained by \ours{} exhibit a higher level of consistency with the ground truth when compared to other methods.
\textcolor{black}{Detailed analysis is presented below.}

\textbf{\textcolor{black}{Results on land use change detection datasets}}. 
Notably, category discrimination approaches yield consistently poor results on DynamicEarthNet and HRSCD, particularly on HRSCD, where the high spatial
resolution introduces complex visual information that makes
it more difficult to distinguish between ground objects. It indicates that semantic segmentation models struggle to accurately capture various semantic information within images, resulting in comparisons with maps that do not effectively reflect real changes. 
As shown in the first scene of~\cref{fig:dynamic earth net comparison}, narrow roads lack distinctive features, making them difficult to segment and resulting in inaccurate label comparisons during change detection.  

Vision discrimination approaches compare pixel similarity directly in the visual feature space, which is generally easier than recognizing the semantic category of each pixel, outperforming category discrimination approaches in most cases. Among them, MapFormer, which leverages a multi-modal feature fusion module to handle cross-modal inputs, generally achieves superior performance compared to bi-temporal change detection methods such as SNUNet and ChangerEx\textcolor{black}{, whose shallow feature fusion or feature exchange mechanisms are tailored for homogeneous image pairs and thus become less effective when applied to heterogeneous map–image inputs}.
However, MapFormer fails to deliver significant performance gains, yielding results comparable to  bi-temporal methods like FHD and ChangeFormer, as its color encoding strategy fails to offer discriminative category
information.

\ours{} leverages language to represent ground objects, which can effectively preserve the high-level categorical information of the map and thus achieves superior performance.

\textbf{\textcolor{black}{Results on building change detection datasets}}.  
\textcolor{black}{We observe that category discrimination approaches achieve better results on the SECOND dataset than on the BANDON dataset.
It can be attributed to the fact that SECOND mainly contains near-nadir imagery, where building locations are well aligned with map annotations, whereas BANDON consists of off-nadir images in which building roofs exhibit noticeable offsets relative to their corresponding locations in the maps, as illustrated in the second scene of~\cref{fig:dynamic earth net comparison}.
Since category discrimination methods first perform semantic segmentation on remote sensing imagery to identify building categories and then compare the predicted labels with map labels on a pixel-by-pixel basis, they are highly sensitive to such spatial misalignment, in addition to the errors introduced by semantic segmentation, resulting in a pronounced performance degradation on the BANDON dataset.}

\textcolor{black}{By comparison, vision discrimination methods and our proposed \ours{} adopt an end-to-end learning paradigm that directly models the correspondence between
maps and images while implicitly capturing the offsets between them. Consequently, these methods are
more robust to the spatial misalignment introduced by off-nadir viewing and maintain relatively stable performance
across both datasets. 
In particular, \ours{} achieves the best performance.
Moreover, by comparing the improvements across different datasets, we can observe that the performance gain is more pronounced on land-use change detection datasets than on building change detection datasets.
It can be considered that the color encoding strategy employed by vision discrimination methods struggles to 
effectively distinguish subtle distinctions among a wide range of categories. 
By leveraging language to bridge the information gap between maps and images, LaVIDE preserves rich categorical semantics from maps and generates more discriminative representations for diverse object categories,} enabling more accurate identification of changed objects and reducing noise from irrelevant changes.

\vspace{-2mm}
\subsection{Ablation Studies}
\subsubsection{Map encoding}
To validate the superiority of map encoding with language, we compare \ours{} with three alternative map encoding strategies, each used to train the proposed network: (1) \ours{}-\textit{Color}, which employs colors to encode maps; (2) \ours{}-\textit{Number}, which encodes maps with numbers; (3) \ours{}-\textit{One-hot}, which represents maps utilizing one-hot encoding.
The results are shown in~\cref{tab:ablation on language}, and we can observe that these three map encoding strategies achieve similar detection performance on both benchmark datasets, as they are too abstract to convey meaningful semantic information, thereby failing to effectively associate high-level category information with low-level visual details.

In contrast, owing to the inherent alignment between language and vision, using language to encode ground objects can effectively bridge the semantic gap between maps and images, resulting in the superior performance of \ours{} compared to the other encoding strategies.
The feature distribution of \ours{}, as shown in~\cref{fig: ablation on language}, exhibits stronger intra-class aggregation and inter-class separation.
Moreover, we observe that \ours{} achieves an average IoU improvement of $15.0\%$ on the DynamicEarthNet dataset, which is significantly higher than the $1.8\%$ gain observed on BANDON. It suggests that the rich semantics encapsulated in language can enhance the association between diverse object categories, thereby offering greater performance advantages in more complex land use change detection scenarios compared to single-category building change detection.

\begin{table}[t]
    \centering
    \caption{Results using different map encoding strategies on DynamicEarthNet and BANDON.}
    \label{tab:ablation on language}
\begin{tabular}{lcccc}
        \toprule
        & \multicolumn{2}{c}{DynamicEarthNet} & \multicolumn{2}{c}{BANDON}\\\cmidrule(lr){2-3}\cmidrule(lr){4-5}
        & F1(\%) & IoU(\%) & F1(\%) & IoU(\%)\\ \midrule
        \ours{}-\textit{Color} & 33.7  & 20.3  & 72.0  & 56.3  \\
        \ours{}-\textit{Number} & 33.6  & 20.2  & 72.3  & 56.6  \\
        \ours{}-\textit{One-hot} & 33.2 & 20.0 & 72.2 & 56.5 \\
        \ours{} & \textbf{37.6}  & \textbf{23.2}  & \textbf{73.0}  & \textbf{57.5}  \\  \bottomrule
        \end{tabular}
    \vspace{-3mm}
\end{table}

\begin{figure}[!t]
    \setlength{\tabcolsep}{0.5mm}
    \renewcommand{\arraystretch}{0.50}
    \newcommand{\firstimage}[1]{dynamic_encoding_#1}
    \centering
    \begin{tabular}{cccc}
	
        \includegraphics[width=0.23\linewidth]{image/exp/\firstimage{color}} &
        \includegraphics[width=0.23\linewidth]{image/exp/\firstimage{number}} &
        \includegraphics[width=0.23\linewidth]{image/exp/\firstimage{onehot}} &
        \includegraphics[width=0.23\linewidth]{image/exp/\firstimage{ours}}
        \\
        \small{\ours{}-\textit{Color}} &
        \small{\ours{}-\textit{Number}} &
	\small{\ours{}-\textit{One-hot}} &
        \small{\ours{}}
    \end{tabular}
	\caption{Visualization of the multi-scale feature $\mathbf{D}$. Red denotes changed samples, whereas green denotes unchanged ones.}
	\vspace{-4mm}
	\label{fig: ablation on language}
\end{figure}

\subsubsection{Prompt design} 
In~\cref{tab:ablation on prompt design}, we evaluate models with different prompt designs for map embedding generation: (1) detailed prompts, which incorporate rich category-specific visual information such as color, texture, and shape; (2) brief prompts, which include only the category name without any additional descriptive information. We adopt three representative prompt templates along with their scene-aware variants incorporating the phrase ``in the scene'' to represent ground objects; 
(3) learnable prompts, which are optimized through supervised learning on downstream tasks. We adopt the standard prompt learning approach~\cite{promptlearning2022} and ours to automatically identify task-specific prompts.

We first compare detailed prompts with brief prompts, which differ primarily in the granularity of semantic content.
We observe that detailed prompts outperform the brief prompts ``A photo of the $C_{i}$.'' and ``The $C_{i}$.'', as the former incorporate visual attributes of categories in remote sensing imagery, which facilitates better alignment with image features.
However, brief prompts that incorporate remote sensing–specific contextual cues, such as ``satellite'' or ``in the scene'', achieve better performance than detailed prompts. 
It suggests that explicitly describing category features in prompts is unnecessary, as the text encoder of LVMs is already aligned with the visual space and can effectively infer visual semantics from concise prompts when appropriately contextualized.

\begin{table}[t]
    \setlength{\tabcolsep}{3pt}
    \centering
    \caption{Results of prompt designs on DynamicEarthNet.}
    \label{tab:ablation on prompt design}
\begin{tabular}{lcc}
        \toprule
        \makecell[l]{Prompts for category information}& F1(\%) & IoU(\%) \\ 
        \midrule
        \textbf{Detailed prompts} \\
        \textit{...$C_{k}$...shades of gray...smooth or uniform texture...} & 32.3 & 19.3 \\
        \midrule
        \textbf{Brief prompts} \\
        $\textit{A photo of the } C_{k} \textit{.}$ & 31.0  & 18.3  \\
        $\textit{A photo of the } C_{k} \textbf{\textit{ in the scene.}}$ & 35.9  & 21.9  \\
        $\textit{The } C_{k} \textit{.}$ & 31.3  & 18.6  \\
        $\textit{The } C_{k} \textbf{\textit{ in the scene.}}$ & 34.4  & 20.8  \\
        $\textit{A satellite photo of the } C_{k} \textit{.}$ & 33.9  & 20.4  \\
        $\textit{A satellite photo of the } C_{k} \textbf{\textit{ in the scene.}}$ & 35.3  & 21.4  \\
        \midrule
        \textbf{Learnable prompts} \\
        $[\bm{t}_{1}]\dots[\bm{t}_{T}]\ C_{k}.$  & 35.1  & 21.3  \\
        $[\bm{t}_{1}]\dots[\bm{t}_{T}]\ C_{k} \textbf{\textit{ in the scene.}}$  &\textbf{37.6}  & \textbf{23.2}    \\ \bottomrule
        \end{tabular}
        \vspace{-6mm}
\end{table}

We then compare learnable prompts with brief prompts.
Due to their ability to automatically adapt to task-specific semantics, learnable prompts in the format ``$[\bm{t}_{1}]\dots[\bm{t}_{T}]\ C_{i}.$'' significantly outperform their handcrafted counterparts such as ``A photo of the $C_i$.'' and ``The $C_i$.''. Our restricted prompt learning strategy further extends this formulation to  ``$[\bm{t}_{1}]\dots[\bm{t}_{T}]\ C_{i} \textbf{\textit{ in the scene.}}$'', which benefits from the integration of image characteristics and facilitates more effective learning of prompts aligned with remote sensing imagery, thereby achieving the best performance.

\subsubsection{Feature enhancement for maps and images}
Our proposed \ours{} network bidirectionally aligns map semantics with image contents through OAEE and KD.
We analyze their effectiveness by comparing variants of \ours{} with and without the OAEE and KD, as shown in~\cref{tab:ablation on object context optimization and knowledge distillation}.
We find that employing either the OAEE or the KD alone results in a certain degree of
performance improvement. The combination of these two strategies leads to a more significant enhancement in performance. 
It indicates that both OAEE and KD contribute meaningfully to improving the alignment between map data and remote sensing imagery.
The visualization results in~\cref{fig:ablation study comparison} also show that incorporating them into \ours{} is better for cross-modal semantic alignment, thus improving change detection performance.

\begin{table}[t]
    \centering
    \caption{Ablation study on OAEE and KD for feature alignment between maps and images on DynamicEarthNet.}
    \label{tab:ablation on object context optimization and knowledge distillation}
\begin{tabular}{c|>{\centering\arraybackslash} p{0.8cm}>{\centering\arraybackslash} p{0.8cm}|>{\centering\arraybackslash} p{1.2cm}>{\centering\arraybackslash} p{1.2cm}}  
\toprule
Method & OAEE & KD & F1(\%) & IoU(\%) \\
\midrule
baseline & & & 33.8 & 20.4 \\
\midrule
\multirow{3}{*}{\ours{}} & \checkmark & & 35.4 & 21.5 \\
 & & \checkmark & 36.0 & 22.0 \\
 & \checkmark & \checkmark & \textbf{37.6} & \textbf{23.2} \\
\bottomrule
    \end{tabular}
    \vspace{-0.4cm}
\end{table}

\begin{figure}[!t]
    \renewcommand{\tabcolsep}{0.2mm}
    \renewcommand{\arraystretch}{0.50}
    \newcommand{\firstimage}[1]{dynamic_module_1_#1}
    \centering
    \begin{tabular}{cccccc}
        \includegraphics[width=0.16\linewidth]{image/exp/\firstimage{map}} &
        \includegraphics[width=0.16\linewidth]{image/exp/\firstimage{image}} &
        \includegraphics[width=0.16\linewidth]{image/exp/\firstimage{baseline}} &
        \includegraphics[width=0.16\linewidth]{image/exp/\firstimage{w_oaee}}&
        \includegraphics[width=0.16\linewidth]{image/exp/\firstimage{w_kd}}&
        \includegraphics[width=0.16\linewidth]{image/exp/\firstimage{w_ours}} \\
	  \small{map} & \small{image} & \small{baseline} & \small{w/ OAEE}& \small{w/ KD}& \small{\ours{}}
    \end{tabular}
	\caption{\textcolor{black}{Qualitative ablation study on OAEE and KD.}}
	\vspace{-6mm}
	\label{fig:ablation study comparison}
\end{figure}

\vspace{-3mm}
\section{Conclusion}
In this paper, we propose a novel map-image change detection algorithm, \ours{}, that leverages language to associate high-level category information with low-level visual details, thus boosting the comparison of maps with images.
It utilizes language to encode maps and aligns map representations with image contents into the feature space of language-vision models for homogenization.
To enhance the effect of feature alignment, we design restricted prompt learning and object-aware embedding enhancement strategies to exploit and enrich map information, making them more consistent with remote sensing image details, thus bridging the information gap for change detection.
Extensive experiments demonstrate that the proposed method can achieve the state-of-the-art on land use change detection (i.e., multi-class) and building change detection (i.e., single-class), outstanding in ensuring the integrity of the change region and suppressing noise.

\ifCLASSOPTIONcaptionsoff
  \newpage
\fi

{\small
\bibliographystyle{IEEEtran}
\bibliography{egbib}
}

\clearpage
\end{document}